\documentclass{article}
\usepackage[preprint]{neurips_2025}
\workshoptitle{In the 39th Conference on Neural Information Processing Systems (NeurIPS 2025) Workshop: Embodied World Models for Decision Making (EWM)}
\usepackage[utf8]{inputenc} 
\usepackage[T1]{fontenc} 
\usepackage[colorlinks=true, allcolors=blue]{hyperref}
\usepackage{url} 
\usepackage{booktabs} 
\usepackage{amsfonts} 
\usepackage{nicefrac} 
\usepackage{microtype} 
\usepackage{xcolor} 
\usepackage{graphicx}
\usepackage{amsmath, amssymb, amsthm}
\title{Generative World Models of Tasks: LLM-Driven Hierarchical Scaffolding for Embodied Agents}
\author{
Brennen Hill\\
Department of Computer Science\\
University of Wisconsin-Madison\\
Madison, WI 53706 \\
\texttt{bahill4@wisc.edu} \\
}
\begin{document}
\maketitle
\begin{abstract}
Recent advances in agent development have focused on scaling model size and raw interaction data, mirroring successes in large language models. However, for complex, long-horizon multi-agent tasks such as robotic soccer, this end-to-end approach often fails due to intractable exploration spaces and sparse rewards. We propose that an effective world model for decision-making must model the world's physics and also its task semantics. A systematic review of 2024 research in low-resource multi-agent soccer reveals a clear trend towards integrating symbolic and hierarchical methods, such as Hierarchical Task Networks (HTNs) and Bayesian Strategy Networks (BSNs), with multi-agent reinforcement learning (MARL). These methods decompose complex goals into manageable subgoals, creating an intrinsic curriculum that shapes agent learning. We formalize this trend into a framework for Hierarchical Task Environments (HTEs), which are essential for bridging the gap between simple, reactive behaviors and sophisticated, strategic team play. Our framework incorporates the use of Large Language Models (LLMs) as generative world models of tasks, capable of dynamically generating this scaffolding. We argue that HTEs provide a mechanism to guide exploration, generate meaningful learning signals, and train agents to internalize hierarchical structure, enabling the development of more capable and general-purpose agents with greater sample efficiency than purely end-to-end approaches.
\end{abstract}

\section{Introduction: Redefining scale for embodied agents}

The paradigm of scaling has demonstrably driven profound advancements in artificial intelligence. In the domain of large language models (LLMs), scaling model parameters, dataset size, and computational resources has unlocked emergent capabilities in reasoning, generation, and comprehension \citep{brown2020language}. A parallel trend is now underway in the development of intelligent agents, where the focus is shifting towards scaling interaction. Environments, rather than passive testbeds, are themselves a crucial dimension of scale. They provide the rich, dynamic data from which agents learn adaptive behaviors, planning, and long-term decision-making. Advancing world models beyond passive prediction and toward active, goal-driven interaction hinges on the quality and structure of this interaction data.

However, as we push agents towards more complex, autonomous, and general-purpose capabilities, a critical bottleneck emerges, particularly in multi-agent systems, which consistently face challenges of non-stationarity and decentralized decision-making \citep{ning2024survey}. In domains characterized by long horizons, combinatorial action spaces, and the need for intricate coordination, such as robotic soccer, the brute-force scaling of interaction data via end-to-end reinforcement learning often proves insufficient. Robotic soccer, as a canonical case study for embodied intelligence, encapsulates these challenges, forcing researchers to develop robust solutions for real-time, strategic cooperation under uncertainty \citep{sammut2010robot}. The challenge is but of \textit{task sparsity}: the sequence of coordinated actions required to achieve a meaningful outcome (e.g., a goal) is so specific and temporally extended that it is almost impossible to discover through random exploration. An agent will not stumble upon a multi-step passing play by chance, regardless of the scale of the simulation. While we ground our analysis in robotic soccer, this challenge of task sparsity is endemic to many complex domains, from collaborative robotic assembly and autonomous driving to automated scientific discovery.

This paper argues for a paradigm shift in how we approach the design of world models and their corresponding environments. We argue that the next crucial frontier is not the scaling of fidelity, size, or agent populations, but the scaling of \textit{structural complexity} within the environment itself. An embodied world model must capture more than just physical dynamics; it must represent the semantic structure of tasks. To address this, we propose a framework for designing Hierarchical Task Environments (HTEs), which embed explicit hierarchical scaffolding into the environment's world model. This scaffolding involves decomposing overarching tasks into a hierarchy of sub-tasks and subgoals, creating an intrinsic curriculum that makes the learning problem tractable. This scaffolding serves as a powerful curriculum, designed to be gradually faded as agents internalize the task structure, forcing them to learn the very hierarchical reasoning skills that were initially externalized in the environment.

To motivate this framework, we first conduct a systematic review of the 2024 literature on multi-agent robotic soccer. Our analysis reveals a clear and compelling trend: the most promising recent approaches are moving away from purely end-to-end MARL and towards hybrid models that integrate symbolic planners and hierarchical reasoning. This paper will demonstrate how these methods implicitly or explicitly leverage task decomposition to overcome the limitations of traditional MARL.
Our contributions are threefold: (1) We identify and analyze the problem of \textit{task sparsity} in complex MARL domains, arguing it is a more fundamental barrier than reward sparsity. (2) We synthesize a clear trend from the 2024 robotic soccer literature, showing a consistent move towards hybrid, hierarchical methods. (3) We propose a novel framework, Hierarchical Task Environments (HTEs), that externalizes task hierarchy into the environment, and we outline a concrete implementation using LLMs as dynamic task planners. We conclude by discussing new evaluation metrics and the implications of this framework for developing generalizable, embodied agents.

\section{Related work} \label{sec:related_work}
Our framework builds upon and seeks to synthesize three established pillars of AI research: Hierarchical Reinforcement Learning (HRL), Curriculum Learning, and the emerging use of LLMs as agents and planners. Our contribution is to advocate for externalizing these principles into the environment's design, creating a more powerful and generalizable learning paradigm.

\textbf{Hierarchical reinforcement learning (HRL)} directly addresses the challenge of long-horizon tasks by imposing a hierarchical structure on an agent's policy. The seminal options framework formalized the concept of temporally extended actions, where a high-level meta-policy selects among a set of options (sub-policies) that execute for multiple timesteps to achieve specific subgoals \citep{sutton1999between}. This temporal abstraction allows agents to reason and plan at a higher level, making credit assignment more tractable and exploration more efficient. This is a form of an internal, learned world model of task structure. Our proposal aims to externalize this hierarchy, embedding the task structure into the environment itself to explicitly guide the learning of such multi-level policies, thereby simplifying the learning problem for the agent.

\textbf{Curriculum learning} is founded on the observation that humans and animals learn more effectively when presented with concepts in a structured, progressive order of increasing difficulty. This principle was formalized for machine learning by \citet{bengio2009curriculum}, demonstrating that training models on a carefully designed curriculum of examples can significantly improve generalization and convergence speed. Many MARL approaches for complex games like soccer employ manually designed curricula, progressing from 1v1 to full team play \citep{baghi2024applying}. We argue that a hierarchically scaffolded environment can generate a curriculum automatically and more dynamically, as an emergent property of the task's compositional structure, moving beyond static, manually-defined stages.

\textbf{LLMs as agents and planners.} Recent work has demonstrated the remarkable potential of LLMs to function as the reasoning core for autonomous agents. This directly aligns with leveraging world knowledge from LLMs. Landmark studies like SayCan have shown that LLMs can decompose high-level natural language instructions into sequences of executable, low-level actions for robotic manipulators, grounding language in physical affordances \citep{ahn2022saycan}. Similarly, Voyager demonstrated an LLM-based agent capable of open-ended exploration and self-directed skill discovery in complex environments like Minecraft, where the LLM itself generates the curriculum by proposing progressively harder tasks \citep{wang2023voyager}. Our proposal leverages this capability, positioning LLMs not just as an agent's brain but as a dynamic, zero-shot environment structurer, a generative world model of tasks that translates high-level goals into the concrete task scaffolding needed for learning.

\section{Problem formulation: Task sparsity in multi-agent learning} \label{sec:problem_formulation}

The term "sparse rewards" has become a ubiquitous explanation for the difficulties encountered in training agents for long-horizon tasks. In robotic soccer, the ultimate reward, scoring a goal, is an infrequent event, making it a challenging signal for credit assignment across a long sequence of actions. While techniques like reward shaping, which provide denser, more localized feedback for actions like moving towards the ball or possessing it, can be effective \citep{li2024marladona, baghi2024applying}, they often treat the symptoms rather than the underlying disease. Beyond simply that the goal is a rare event, the foundational problem is that the path to that goal requires a long, temporally extended, and specific sequence of coordinated sub-tasks. We identify the core challenge as one of \textit{task decomposition}.

\subsection{Task sparsity in multi-agent coordination}

Consider the sequence of actions required for a successful offensive play in 3v3 soccer: Agent 1 must first gain possession of the ball. It must then assess the positions of its teammates and opponents. It might then decide to dribble past an immediate defender. Subsequently, it must execute a pass to Agent 2, who is moving into an open position. Agent 2 must receive the pass, orient towards the goal, and perhaps take a shot. Each of these steps is a subgoal, and failure at any point breaks the chain.

An end-to-end MARL agent, operating with a flat action space and a single terminal reward, faces a combinatorial explosion. If each of $N$ agents has an action space of size $|A|$, the joint action space at each timestep is $|A|^N$. Over a horizon of $T$ steps, the number of possible trajectories is astronomical. The probability of a random or semi-random exploration policy discovering this entire successful sequence is infinitesimally small. This is the problem of \textit{task sparsity}: the density of successful, goal-achieving trajectories in the vast space of all possible trajectories is effectively zero. Simply providing a denser reward for getting closer to the goal does not teach the agent the crucial concepts of passing, creating space, or defending, the abstract building blocks of strategy.

\subsection{The inefficiency of brute-force scaling}

The prevailing approach in machine learning has been to overcome such challenges with scale. However, the nature of multi-agent coordination makes this particularly inefficient. As noted in the analysis of a modified TiZero framework, MARL methods can learn meaningful strategies but still struggle with sample inefficiency and slow learning progress \citep{baghi2024applying}. The computational cost scales exponentially with the number of agents and the complexity of the required coordination.

Furthermore, a focus on extreme computational scale can make research inaccessible and sideline promising alternative paradigms. Our analysis of the 2024 soccer literature (see Section \ref{sec:analysis}) was intentionally focused on low-resource approaches that control agents at a high level (e.g., target positions, kick actions) rather than low-level joint angles. This decision was informed by the observation that some of the most complex end-to-end systems, such as those that learn directly from egocentric vision \citep{tirumala2024learning}, require a scale of resources that places them beyond the reach of most academic labs. These represent a brute-force approach to a problem that may be better solved with structural priors. The goal should be to find more efficient paths to intelligence, instead of the most computationally expensive ones.

\section{Analysis of hierarchical methods in multi-agent robotic soccer} \label{sec:analysis}

Our review of the 2024 literature uncovers a significant trend: a move away from monolithic, end-to-end MARL and towards hybrid architectures that explicitly integrate hierarchical and symbolic reasoning. These approaches directly confront the problem of task decomposition, providing a structured framework that guides learning and makes complex, long-horizon coordination tractable. A summary of the most relevant papers is presented in Table \ref{tab:literature_summary}.

\subsection{Methodology of review}

To ground our proposal in empirical evidence, we conducted a systematic survey of academic papers published in 2024 that addressed multi-agent robotic soccer. Our search criteria targeted papers in reputable journals and conference proceedings containing the terms "multi-agent" and "soccer." We applied several filters to align the review with the problem of strategic, low-resource learning:

\begin{enumerate}
\item \textbf{Domain focus:} We excluded papers where soccer was merely one of several benchmark environments for a general MARL algorithm, focusing instead on work where soccer was the primary domain of investigation. This focus on a competition-driven domain is deliberate; major robotics competitions are consistently identified as primary drivers of innovation, providing standardized benchmarks and forcing researchers to integrate solutions into effective, holistic systems \citep{brancaliao2022systematic, sammut2010robot}.
\item \textbf{Abstraction level:} We excluded papers that focused on low-level control, such as learning individual robot joint angles for walking or kicking, to concentrate on the strategic and tactical layers of decision-making relevant to long-horizon planning.
\item \textbf{Resource accessibility:} We prioritized methods that demonstrated effectiveness without relying on massive, proprietary computational infrastructure, aiming to identify scalable principles rather than just scalable implementations.
\end{enumerate}
This curated selection of recent work forms the basis for the analysis presented in the next section.

\begin{table}[h!]
\centering
\caption{Summary of key 2024 multi-agent soccer literature demonstrating a trend towards hierarchical and structured approaches.}
\label{tab:literature_summary}
\resizebox{\textwidth}{!}{%
\begin{tabular}{@{}llll@{}}
\toprule
\textbf{Paper} & \textbf{Core method} & \textbf{Key contribution} & \textbf{Relation to hierarchy/structure} \\ \midrule
\citet{li2024marladona} & MARLadona (MARL) & Dense rewards + curriculum for team play. & Explicit, hand-crafted curriculum (1v1, 2v2). \\
 & & & Reward shaping provides implicit goal structure. \\ \\
\citet{baghi2024applying} & Modified TiZero (MARL) & Explores intrinsic rewards in a curriculum. & Heavily relies on a multi-stage, explicit curriculum. \\ \\
\citet{mu2024hsmarl} & HS-MARL & Integrates HTN planner with MARL. & Explicit hierarchical task decomposition via symbolic planner. \\
 & & Meta-controller assigns subgoals to agents. & Symbolic options create a structured, high-level action space. \\ \\
\citet{azarkasb2024eligibility} & Eligibility Traces & Learns a knowledge base of successful paths. & Implicit structural memory via traces; multi-stage training curriculum. \\ \\
\citet{li2024selfattention} & ADA (Advice Distillation) & Teacher agent provides advice to student. & Social hierarchy (teacher-student) structures learning process. \\ \\
\citet{zhao2024mimic} & MCRL (Mimic-to-Counteract) & Two-stage curriculum: mimic expert, then counteract. & Structures training through distinct, high-level strategic goals. \\ \\
\citet{Yang2024Bayesian} & BSN (Bayesian Strategy Nets) & Decomposes policy into sub-policies using a BSN. & Explicit policy decomposition into simpler, coordinated components. \\ \bottomrule
\end{tabular}%
}
\end{table}

\subsection{Explicit task decomposition with symbolic planners} \label{sec:explicit_task}

The most compelling evidence for our proposal comes from the work of \citet{mu2024hsmarl}, who introduce the HS-MARL framework. This work stands as a direct confirmation that task decomposition is the central challenge. Instead of relying on RL to discover high-level strategy from scratch, HS-MARL integrates a classical AI planner, a Hierarchical Task Network (HTN) planner called pyHIPOP+, into the MARL loop.

The core idea is to use the HTN planner to decompose the overarching goal (e.g., score a goal) into a logically sound sequence of executable subgoals. This plan is generated using a symbolic representation of the world state and a domain description written in the Hierarchical Domain Definition Language (HDDL). The framework constructs a hierarchical state-space tree where subgoals are arranged based on their logical and temporal proximity to the final goal. For instance, the goal ScoreGoal might be decomposed into the subtasks AcquireBall, MoveToShootingPosition, and ExecuteShot. A meta-controller then uses this symbolic plan to guide the low-level MARL agents. It assigns symbolic options, policies trained to achieve specific subgoals, to the agents. This architecture directly addresses task sparsity by providing a high-level scaffold for the agents' behavior. The RL component is no longer responsible for discovering the entire strategic sequence but is instead focused on learning how to execute the sub-tasks prescribed by the planner. The authors explicitly note that this integration is designed to tackle complex environments with sparse rewards, demonstrating significantly improved performance and sample efficiency.

This trend is reinforced by related work from some of the same authors on Bayesian Strategy Networks (BSN), which aims to separate an intricate policy into several simple sub-policies and organize their relationships \citep{Yang2024Bayesian}. A BSN models the conditional dependencies between sub-policies (e.g., a shoot sub-policy is only relevant if the "possess ball" condition is met), effectively decomposing the global policy into a structured, interpretable graph. While distinct in its probabilistic approach, the underlying philosophy is identical: complex behavior emerges from the composition of simpler, learned components.

\subsection{Hierarchical structure as an implicit curriculum}

A key benefit of embedding hierarchical structure is the natural emergence of a curriculum, which guides agent learning from simple to complex behaviors. Manually designing such curricula is a known effective technique, but it is labor-intensive and difficult to scale. For example, the TiZero framework for soccer relies on a curriculum of manually designed scenarios of increasing difficulty \citep{baghi2024applying}. An agent must achieve a certain win rate in a simple 1v1 scenario before progressing to a more complex 2v2 scenario, and so on.

The hierarchical scaffolding approach automates this process. In HS-MARL, the state-space tree generated by the HTN planner forms a natural curriculum. Agents first learn policies for subgoals at the leaves of the decomposition tree (e.g., \texttt{move\_to\_ball}), as these are prerequisites for higher-level subgoals (e.g., \texttt{pass\_to\_teammate}). The structure of the task itself dictates the learning progression, making the curriculum an emergent property of the environment's hierarchical design rather than a manually engineered artifact.

A different, yet philosophically similar, approach to emergent curricula can be seen in the work of \citet{azarkasb2024eligibility}. Their method uses eligibility traces to build a knowledge base of successful trajectories. Critically, their training process is multi-staged: first, the agent learns optimal paths to the ball and goal in an obstacle-free environment. Only after mastering this foundational skill are obstacles (i.e., opponent robots) introduced. This two-stage process constitutes a simple but effective curriculum that emerges directly from the incremental layering of task complexity. The stored eligibility traces function as a learned library of successful sub-plans, akin to the symbolic options in HS-MARL.

\subsection{Abstraction for efficient and interpretable learning}

A third major advantage of the hierarchical approach is that it facilitates learning in a more abstract, and therefore more tractable, state-action space. By decomposing the problem, we can define high-level actions and representations that hide irrelevant low-level details, drastically reducing the exploration space for the learning algorithm.

The HS-MARL framework exemplifies this principle. The meta-controller operates in the symbolic space of subgoals and options, while the low-level agents execute these options in the continuous state space. This separation of concerns allows the strategic layer (the planner and meta-controller) to reason over long horizons without getting bogged down in the minutiae of motor control. As \citet{mu2024hsmarl} note, a key benefit of this architecture is enhanced interpretability. One can inspect the symbolic plan generated by the HTN to understand the agent team's high-level strategy, a task that is nearly impossible with a monolithic neural network policy.

This move towards abstraction is a field-wide trend. Our decision to exclude joint-level control from our review reflects this shift. Modern MARL frameworks for soccer, like MARLadona \citep{li2024marladona}, operate with abstract actions like move, turn, and kick, rather than controlling individual motor torques. Similarly, methods focused on high-level strategy, such as advice distillation from a teacher agent \citep{li2024selfattention} or policy distillation for countering opponents \citep{zhao2024mimic}, presuppose an abstract, strategic layer of interaction. These methods work because they operate on a representation of the problem that is already simplified and structured. Hierarchical scaffolding is the logical next step: making that structure an explicit and dynamic component of the learning process itself.

\section{HTE: A framework for Hierarchical Task Environments} \label{sec:ste_framework}

Based on our analysis, we propose a new framework for environment design to address the task sparsity challenge. The dominant paradigm of creating high-fidelity, flat simulators and challenging agents to learn from scratch is fundamentally inefficient for teaching strategy. We must shift our focus to designing world models and environments that are themselves structured for learning.

\subsection{From flat physics simulators to Hierarchical Task Environments}
Many current multi-agent benchmarks, such as Google Research Football \citep{kurach2020google} or the Isaac Gym-based environment used by MARLadona \citep{li2024marladona}, are best described as sophisticated physics simulators. They provide the laws of the world and a sparse terminal goal, leaving the entire burden of task decomposition and strategy discovery to the learning algorithm. While valuable for testing raw learning capabilities, their world models are incomplete for decision-making because they lack a model of task semantics.

We propose that the next generation of benchmarks should be architected as \textit{Hierarchical Task Environments} (HTEs). This means moving beyond physics and providing explicit support for hierarchical task definition and decomposition as a core feature of the environment's world model itself. The environment should be a place where an agent learns structure instead of just be a place where an agent acts.

\subsection{Framework components} \label{sec:framework_components}

A scaffolded environment would possess several key features that distinguish it from a traditional simulator:

\begin{enumerate}
\item \textbf{A hierarchical task API:} The environment would expose an API for defining tasks and sub-tasks, along with their dependencies and termination conditions. This would be analogous to the HDDL domains used by \citet{mu2024hsmarl}, but integrated directly into the environment's state management. For instance, a researcher could define a \texttt{PassingPlay} task composed of the sub-tasks \texttt{passer\_moves\_to\_ball}, \texttt{receiver\_moves\_to\_open\_space}, and \texttt{passer\_kicks\_to\_receiver}. (See Appendix \ref{sec:appendix_a} for a minimal specification).

\item \textbf{Layered action spaces:} Agents could interact with the environment at multiple levels of abstraction, from low-level continuous commands (e.g., \texttt{set\_velocity(x, y)}) to high-level symbolic options (e.g., \texttt{execute\_option(pass\_to\_teammate\_3)}). This directly supports the development of hierarchical policy architectures.

\item \textbf{Intrinsic rewards for subgoals:} The environment itself would manage the reward signals associated with the task hierarchy. Upon successful completion of a defined subgoal, the environment would issue an intrinsic reward to the relevant agent(s). This codifies reward shaping directly into the task definition, providing dense, meaningful learning signals that are perfectly aligned with the task's compositional structure.

\item \textbf{Procedural curriculum generation:} With a compositional task definition system, the environment could procedurally generate curricula, starting with foundational sub-tasks and gradually combining them into more complex, longer-horizon challenges as agents demonstrate mastery.

\item \textbf{Built-in support for compositional evaluation:} Such environments necessitate new evaluation metrics beyond simple win rates. We propose:
    \begin{itemize}
        \item \textbf{Compositional generalization score:} Measures an agent's zero-shot or few-shot performance on a novel, complex task composed of sub-tasks it has already mastered individually. \textbf{The protocol would involve training agents on a set of 'base' sub-tasks (e.g., \texttt{dribble}, \texttt{pass}) and then testing them on a held-out compositional task (e.g., \texttt{give-and-go}) without further training, comparing their success rate to a flat baseline.}
        \item \textbf{Curriculum efficiency:} Quantifies the reduction in sample complexity ($1/N_{samples}$) or wall-clock time required to reach a target performance level in the scaffolded environment compared to a flat baseline. This is measured as the ratio of samples-to-convergence ($N_{flat} / N_{scaffolded}$) where the flat baseline is a state-of-the-art MARL agent trained in the same environment with only the terminal goal reward (e.g., scoring).
        \item \textbf{Scaffolding brittleness index:} Assesses the performance degradation when the provided task hierarchy is intentionally made sub-optimal. We propose measuring this by comparing the final performance of an agent trained with the full scaffold ($P_{full}$) to one trained with a faded scaffold that has been partially removed ($P_{faded}$), and to an agent trained on a sub-optimal or intentionally wrong hierarchy ($P_{wrong}$). A robust agent would show minimal degradation, indicating it has internalized the strategy rather than simply overfitting to the scaffold.
    \end{itemize}

\end{enumerate}

\subsection{Dynamic scaffolding with LLM-based task planners}

While the HTN-based approach in HS-MARL provides a powerful proof-of-concept, its reliance on manually crafted, domain-specific HDDL files represents a significant engineering bottleneck. Recent work has already demonstrated the power of LLMs to decompose high-level instructions into executable plans for embodied agents \citep{ahn2022saycan, wang2023voyager}. We propose that this principle can be adapted to dynamically structure the learning environment itself, transforming static scaffolding into a flexible, language-driven curriculum. The LLM becomes a \textit{generative world model of tasks}.

In this paradigm, an environment would not have a fixed task hierarchy. Instead, a high-level goal would be specified in natural language (e.g., "Execute a give-and-go play with the forward on the right wing"). An integrated LLM would then perform common-sense reasoning to decompose this instruction into a plausible, ordered sequence of symbolic subgoals: [1. Player A dribbles towards Player B], [2. Player A passes to Player B], [3. Player A runs into open space past defender], [4. Player B returns pass to Player A]. This sequence would be passed to the environment's task API (see Appendix \ref{sec:appendix_a}), dynamically configuring the intrinsic reward function and success conditions for that specific episode. This approach offers several transformative advantages:

\begin{itemize}
\item \textbf{Scalability and flexibility:} It eliminates the need for human experts to write complex domain description files for every conceivable strategy. New tasks and curricula can be generated simply by writing new prompts.
\item \textbf{Instruction-following agents:} It provides a natural framework for training agents that can follow complex, high-level instructions, a cornerstone of general-purpose agency and a key topic in VLA models.
\item \textbf{Dynamic adaptation:} The LLM planner could potentially re-plan mid-episode based on new state information, providing a dynamic scaffold that adapts to the unfolding situation on the field. This directly addresses a key limitation of rigid, pre-computed HTN plans.
\end{itemize}

LLM-generated plans introduce challenges of logical consistency and physical feasibility, as noted by \citet{ahn2022saycan}. A robust implementation would require a \textit{verification layer} to augment the LLM. This blueprint would involve (a) symbolic constraint checking (e.g., verifying that \texttt{is\_possessing(ball)} is true before dispatching a \texttt{pass} subgoal) and (b) physical feasibility queries to the world model (e.g., "Is the path for this pass clear?"). If a plan fails verification, the system could \textit{re-prompt} the LLM with the specific error, or \textit{rollback} to a previously validated state, creating a robust planning-execution loop.

\subsection{Implications for agent architectures and Sim2Real transfer}

Shifting the focus to scaffolded environments would catalyze a corresponding shift in agent architectures and accelerate research in several key areas. Architectures with high-level meta-policies that select among low-level sub-policies, as formalized in the options framework \citep{sutton1999between}, would be a natural fit. This also provides a powerful grounding for hybrid, neuro-symbolic agents that combine neural pattern recognition with logical reasoning. Furthermore, an agent that has learned a policy for a "move to open space" sub-task could more easily reuse that skill when it appears as part of a new, more complex task, making scaffolded environments ideal testbeds for studying generalization through composition.

Crucially, this paradigm offers a more tractable path for Sim2Real transfer. Hierarchical policies are inherently better suited for transfer because the domain shift is often concentrated at the lowest levels. The high-level strategic layer (e.g., deciding when to pass) is often robust to the differences between simulation and reality, while the reality gap primarily affects the low-level motor policies (e.g., the precise joint torques needed to execute the pass) \citep{kober2013reinforcement}. This decomposition allows researchers to focus domain adaptation efforts on a smaller set of well-defined sub-tasks, making the Sim2Real problem more tractable for embodied agents.

\section{Discussion and future work} \label{sec:discussion}

While promising, the paradigm of scaffolded environments presents its own set of research challenges that must be addressed.

\subsection{The scaffolding design problem}

A primary challenge is the origin of the hierarchical task structure itself. If the scaffolds must be meticulously hand-crafted by human experts, we have merely shifted the burden from the learning algorithm to the environment designer. This creates a bottleneck and may imbue the agent with the designer's own biases. The aforementioned use of LLMs as zero-shot planners is a promising mitigation strategy, but it introduces its own challenges, such as ensuring the logical consistency and physical feasibility of LLM-generated plans. Future work could explore methods for agents to learn the task hierarchy itself, perhaps by identifying bottleneck states in their exploration or by abstracting common sub-trajectories into reusable skills. To mitigate this initial "operator burden," a practical bootstrapping recipe could involve learning from demonstration. An initial library of sub-tasks and options could be learned by observing human expert data, with an LLM then labeling and abstracting these demonstrated trajectories into the symbolic task hierarchy. This human-in-the-loop approach seeds the system with a library of primitives, which can then be composed into new tasks.

\subsection{The risk of over-constraining discovery}
A critical concern is whether providing explicit structure could stifle discovery or reduce generality. An agent trained in a scaffolded environment might fail in a flat one, having used the hierarchy as a crutch rather than a learning aid.

We argue that this perspective misinterprets the purpose of the scaffold. The scaffold is a curriculum rather than the final environment. Instead of simply solving the scaffolded task, the goal is to use the scaffold to shape the agent's internal world model to represent task hierarchies. The training process must therefore include scaffold fading: as the agent masters sub-tasks, the explicit environmental rewards and API-level guidance are gradually removed. This process forces the agent to \textit{internalize} the hierarchical policy, analogous to how a student, after learning with a tutor, must eventually take the test alone. The fading process itself becomes a key part of the curriculum, training the agent to rely on its own learned, internal task-decomposition model. The proposed Scaffolding Brittleness Index is designed precisely to measure an agent's success in this internalization and ensure it is not overfitted to the training wheels.
\subsection{The grounding problem}

The effective use of symbolic subgoals (whether from an HTN or an LLM) depends on the ability to connect them to the low-level, sub-symbolic state of the environment. This is the classic symbol grounding problem \citep{harnad1990symbol}. For example, how does an agent robustly translate the symbolic subgoal "move to open space" into a concrete target $(x, y)$ coordinate in a dynamic environment with moving teammates and opponents? Solving this requires robust perception models that can map between the continuous state space (e.g., from vision) and a discrete, symbolic representation. Developing such models that are both accurate and learnable is a central challenge for neuro-symbolic agent architectures and is directly related to VLA models that must align language with perception and action.
\subsection{Generality beyond robotic soccer}

While we have grounded our proposal in soccer, this environment-as-scaffold paradigm transfers directly to other complex, long-horizon domains. For instance, in collaborative robotic assembly, the Task API could define a hierarchy of \texttt{install\_component} tasks, with subgoals for \texttt{fetch\_part}, \texttt{align\_part}, and \texttt{secure\_part}. In autonomous driving, a high-level instruction like "merge onto the highway" could be dynamically decomposed by the environment into a curriculum of subgoals: \texttt{match\_speed\_with\_traffic}, \texttt{find\_safe\_gap}, \texttt{execute\_lane\_change}, and \texttt{establish\_follow\_distance}. In both cases, the environment provides the structural priors needed to make an intractable, long-horizon problem solvable.
\subsection{Interaction with MARL pathologies}

A scaffolded environment does not eliminate fundamental MARL challenges such as non-stationarity or partial observability, but it does reframe them. By assigning specific, independent sub-tasks to different agents, the hierarchy can stabilize the learning problem and provide a structured solution to credit assignment. However, it also raises new questions: how should an agent behave if its partner, operating under partial observability, fails at a prerequisite sub-task? Future work could explore how these explicit task dependencies interact with the implicit social dynamics and decentralized nature of MARL.
\section{Conclusion}

The pursuit of more capable and autonomous agents has led the research community to recognize the critical role of environments in shaping intelligence. The current emphasis on scaling the fidelity and size of these environments, while valuable, overlooks a more crucial dimension: structural complexity. This paper analyzed the limitations of end-to-end MARL in complex, long-horizon domains, identifying \textit{task sparsity} as a fundamental barrier. An embodied world model for decision-making must therefore model both the physics of the world and the structure of tasks within it.

Our review of the 2024 literature in multi-agent soccer revealed a clear trend towards hybrid systems that integrate symbolic planning and hierarchical decomposition with MARL. These methods succeed precisely because they provide the structural priors necessary to make the problem of strategic coordination tractable.

Building on this analysis, we proposed the Structured Task Environment (HTE) framework, which externalizes task hierarchy into the world model itself. We further proposed a novel implementation using LLMs as dynamic, generative world models of tasks to create this structure on the fly. This paradigm directly enables research into long-horizon planning, compositional generalization, and language-guided decision-making. To scale agents to complex, strategic domains, we must first scale the environments to support complex, strategic learning. Future work will focus on implementing the proposed LLM-based task API and validating the framework's impact on sample efficiency and generalization in benchmark domains.

\bibliographystyle{plainnat}
\bibliography{references}

\begin{thebibliography}{19}
\providecommand{\natexlab}[1]{#1}
\providecommand{\url}[1]{\texttt{#1}}
\expandafter\ifx\csname urlstyle\endcsname\relax
  \providecommand{\doi}[1]{doi: #1}\else
  \providecommand{\doi}{doi: \begingroup \urlstyle{rm}\Url}\fi

\bibitem[Ahn et~al.(2022)Ahn, Brohan, Brown, Chebotar, Cortes, David, Finn, Fu, Gopalakrishnan, Hausman, Herzog, Ho, Hsu, Ibarz, Ichter, Irpan, Jang, Jauregui~Ruano, Jeffrey, Jesmonth, Joshi, Julian, Kalashnikov, Kuang, Lee, Levine, Lu, Luu, Parada, Pastor, Quiambao, Rao, Rettinghouse, Reyes, Sermanet, Sievers, Tan, Toshev, Vanhoucke, Xia, Xiao, Xu, Xu, Yan, and Zeng]{ahn2022saycan}
Michael Ahn, Anthony Brohan, Noah Brown, Yevgen Chebotar, Omar Cortes, Byron David, Chelsea Finn, Chuyuan Fu, Keerthana Gopalakrishnan, Karol Hausman, Alex Herzog, Daniel Ho, Jasmine Hsu, Julian Ibarz, Brian Ichter, Alex Irpan, Eric Jang, Rosario Jauregui~Ruano, Kyle Jeffrey, Sally Jesmonth, Nikhil~J Joshi, Ryan Julian, Dmitry Kalashnikov, Yuheng Kuang, Kuang-Huei Lee, Sergey Levine, Yao Lu, Linda Luu, Carolina Parada, Peter Pastor, Jornell Quiambao, Kanishka Rao, Jarek Rettinghouse, Diego Reyes, Pierre Sermanet, Nicolas Sievers, Clayton Tan, Alexander Toshev, Vincent Vanhoucke, Fei Xia, Ted Xiao, Peng Xu, Sichun Xu, Mengyuan Yan, and Andy Zeng.
\newblock Do as i can, not as i say: Grounding language in robotic affordances.
\newblock \emph{arXiv preprint arXiv:2204.01691}, 2022.

\bibitem[Azarkasb and Khasteh(2024)]{azarkasb2024eligibility}
Seyed~Omid Azarkasb and Seyed~Hossein Khasteh.
\newblock {Eligibility Traces in an Autonomous Soccer Robot with Obstacle Avoidance and Navigation Policy}.
\newblock \emph{Journal of Intelligent \& Robotic Systems}, 109\penalty0 (1):\penalty0 1--20, 2024.

\bibitem[Baghi(2024)]{baghi2024applying}
Amir~Masoud Baghi.
\newblock \emph{{Applying Multi-Agent Reinforcement Learning as Game-AI in Football-like Environments}}.
\newblock PhD thesis, KTH Royal Institute of Technology, 2024.

\bibitem[Bengio et~al.(2009)Bengio, Louradour, Collobert, and Weston]{bengio2009curriculum}
Yoshua Bengio, J{\'e}r{\^o}me Louradour, Ronan Collobert, and Jason Weston.
\newblock Curriculum learning.
\newblock In \emph{Proceedings of the 26th annual international conference on machine learning}, pages 41--48, 2009.

\bibitem[Brancalião et~al.(2022)Brancalião, Gonçalves, Conde, and Costa]{brancaliao2022systematic}
Laiany Brancalião, José Gonçalves, Miguel~Á. Conde, and Paulo Costa.
\newblock {Systematic Mapping Literature Review of Mobile Robotics Competitions}.
\newblock \emph{Sensors}, 22\penalty0 (6):\penalty0 2160, mar 2022.
\newblock \doi{10.3390/s22062160}.
\newblock URL \url{https://doi.org/10.3390/s22062160}.

\bibitem[Brown et~al.(2020)Brown, Mann, Ryder, and et~al.]{brown2020language}
Tom~B Brown, Benjamin Mann, Nick Ryder, and et~al.
\newblock Language models are few-shot learners.
\newblock \emph{Advances in neural information processing systems}, 33:\penalty0 1877--1901, 2020.

\bibitem[Harnad(1990)]{harnad1990symbol}
Stevan Harnad.
\newblock The symbol grounding problem.
\newblock \emph{Physica D: Nonlinear Phenomena}, 42\penalty0 (1-3):\penalty0 335--346, 1990.

\bibitem[Kober et~al.(2013)Kober, Bagnell, and Peters]{kober2013reinforcement}
Jens Kober, J~Andrew Bagnell, and Jan Peters.
\newblock Reinforcement learning in robotics: A survey.
\newblock \emph{The International Journal of Robotics Research}, 32\penalty0 (11):\penalty0 1238--1274, 2013.

\bibitem[Kurach et~al.(2020)Kurach, Raichuk, Sta{\'n}czyk, and et~al.]{kurach2020google}
Karol Kurach, Anton Raichuk, Piotr Sta{\'n}czyk, and et~al.
\newblock Google research football: A novel reinforcement learning environment.
\newblock In \emph{Proceedings of the AAAI Conference on Artificial Intelligence}, volume~34, pages 4501--4510, 2020.

\bibitem[Li et~al.(2024{\natexlab{a}})Li, Zhou, Hou, Zhou, Ge, and Feng]{li2024selfattention}
Yang Li, Sihan Zhou, Yaqing Hou, Liran Zhou, Hongwei Ge, and Liang Feng.
\newblock {Self-Attention Guided Advice Distillation in Multi-Agent Deep Reinforcement Learning}.
\newblock \emph{IEEE Transactions on Neural Networks and Learning Systems}, 2024{\natexlab{a}}.

\bibitem[Li et~al.(2024{\natexlab{b}})Li, Bjelonic, Klemm, and Hutter]{li2024marladona}
Zichong Li, Filip Bjelonic, Victor Klemm, and Marco Hutter.
\newblock {MARLadona - Towards Cooperative Team Play Using Multi-Agent Reinforcement Learning}.
\newblock \emph{arXiv preprint arXiv:2401.12345}, 2024{\natexlab{b}}.

\bibitem[Mu et~al.(2024)Mu, Zhuo, Chen, Zhang, Yu, and Hao]{mu2024hsmarl}
Xuechen Mu, Hankz~Hankui Zhuo, Chen Chen, Kai Zhang, Chao Yu, and Jianye Hao.
\newblock {Hierarchical Task Network-Enhanced Multi-Agent Reinforcement Learning: Toward Efficient Cooperative Strategies}.
\newblock In \emph{Proceedings of the AAAI Conference on Artificial Intelligence}, volume~38, pages 16853--16861, 2024.

\bibitem[Ning and Xie(2024)]{ning2024survey}
Zepeng Ning and Lihua Xie.
\newblock {A Survey on Multi-Agent Reinforcement Learning and its Application}.
\newblock \emph{Engineering Applications of Artificial Intelligence}, 127:\penalty0 107335, 2024.

\bibitem[Sammut(2010)]{sammut2010robot}
Claude Sammut.
\newblock Robot soccer.
\newblock \emph{Wiley Interdisciplinary Reviews: Cognitive Science}, 1\penalty0 (6):\penalty0 824--833, 2010.

\bibitem[Sutton et~al.(1999)Sutton, Precup, and Singh]{sutton1999between}
Richard~S Sutton, Doina Precup, and Satinder Singh.
\newblock Between mdps and semi-mdps: A framework for temporal abstraction in reinforcement learning.
\newblock \emph{Artificial intelligence}, 112\penalty0 (1-2):\penalty0 181--211, 1999.

\bibitem[Tirumala et~al.(2024)Tirumala, Wulfmeier, Moran, Huang, Humplik, Lever, Haarnoja, Hasenclever, Byravan, Batchelor, Sreendra, Patel, Gwira, Nori, Riedmiller, and Heess]{tirumala2024learning}
Dhruva Tirumala, Markus Wulfmeier, Ben Moran, Sandy Huang, Jan Humplik, Guy Lever, Tuomas Haarnoja, Leonard Hasenclever, Arunkumar Byravan, Nathan Batchelor, Neil Sreendra, Kushal Patel, Marlon Gwira, Francesco Nori, Martin Riedmiller, and Nicolas Heess.
\newblock {Learning Robot Soccer from Egocentric Vision with Deep Reinforcement Learning}.
\newblock \emph{Robotics: Science and Systems}, 2024.

\bibitem[Wang et~al.(2023)Wang, Xie, Jiang, and et~al.]{wang2023voyager}
Guanzhi Wang, Yuqi Xie, Yunfan Jiang, and et~al.
\newblock Voyager: An open-ended embodied agent with large language models.
\newblock \emph{arXiv preprint arXiv:2305.16291}, 2023.

\bibitem[Yang and Parasuraman(2024)]{Yang2024Bayesian}
Qin Yang and Ramviyas Parasuraman.
\newblock Bayesian strategy networks based soft actor-critic learning.
\newblock \emph{ACM Trans. Intell. Syst. Technol.}, 15, mar 2024.
\newblock \doi{10.1145/3643862}.

\bibitem[Zhao et~al.(2024)Zhao, Lin, Zhang, Li, Zhou, and Sun]{zhao2024mimic}
Junjie Zhao, Jiangwen Lin, Xinyan Zhang, Yuanbai Li, Xianzhong Zhou, and Yuxiang Sun.
\newblock {From Mimic to Counteract: a Two-Stage Reinforcement Learning Algorithm for Google Research Football}.
\newblock \emph{International Conference on Autonomous Agents and Multiagent Systems (AAMAS)}, 2024.

\end{thebibliography}

\appendix
\section{Technical appendices and supplementary material}

\label{sec:appendix_a}

To make our proposal concrete, we outline a minimal specification for the Hierarchical Task API described in Section \ref{sec:framework_components}.
\subsection{A.1 Task schema}
A task can be defined by a data structure that specifies its components. A planner (like an HTN or LLM) would populate these structures, which the environment would then use to manage the learning process.
\begin{verbatim}
class HierarchicalTask:
def init(self, name, parent_task, subtasks):
self.name = name # e.g., "GiveAndGo"
self.parent = parent_task # Task object or None
self.subtasks = subtasks # Ordered list of Task objects
self.status = "inactive" # inactive, active, completed, failed
def check_preconditions(self, world_state):
    # Logic to check if this task can start
    # e.g., for "PassToTeammate", check "agent_has_ball"
    pass

def check_termination(self, world_state):
    # Logic to check if this task is done
    # e.g., for "PassToTeammate", check "teammate_has_ball"
    pass

def get_intrinsic_reward(self, world_state):
    # Return reward if check_termination() is true
    pass


\end{verbatim}
\subsection{A.2 Example: HDDL-to-API mapping}
As noted in Section \ref{sec:explicit_task}, an HDDL planner can generate the task structure. A high-level HTN method for a (ScoreGoal) task might be:
\begin{verbatim}
(:method (ScoreGoal)
:subtasks (and
(AcquireBall ?agent1)
(DribblePastOpponent ?agent1 ?opponent)
(PassToTeammate ?agent1 ?agent2)
(ShootAtGoal ?agent2)
)
)
\end{verbatim}
This would map to our API as follows:
\begin{verbatim}
Create sub-tasks (as "leaf" nodes)
task_acquire = HierarchicalTask("AcquireBall", ...)
task_dribble = HierarchicalTask("DribblePastOpponent", ...)
task_pass = HierarchicalTask("PassToTeammate", ...)
task_shoot = HierarchicalTask("ShootAtGoal", ...)
Create the parent "composite" task
task_score = HierarchicalTask(
name="ScoreGoal",
parent=None,
subtasks=[task_acquire, task_dribble, task_pass, task_shoot]
)
\end{verbatim}
\subsection{A.3 Standardized soccer task definitions}
Here are two standardized task definitions using this structure.

\textbf{1. Task: GainPossession}
\begin{itemize}
\item \textbf{Description:} A single agent must gain control of the ball.
\item \textbf{name}: "GainPossession"
\item \textbf{subtasks}: [ \texttt{MoveToBall}, \texttt{ControlBall} ]
\item \textbf{preconditions(MoveToBall)}: \texttt{ball.is\_loose()}
\item \textbf{termination(MoveToBall)}: $distance(agent, ball) < 0.5m$
\item \textbf{termination(ControlBall)}: \texttt{agent.has\_ball == True}
\item \textbf{get\_intrinsic\_reward()}: $+0.5$ on \texttt{MoveToBall} completion, $+1.0$ on \texttt{ControlBall} completion.
\end{itemize}
\textbf{2. Task: GiveAndGo} (Multi-agent)
\begin{itemize}
\item \textbf{Description:} Agent A passes to Agent B, runs past a defender, and receives a return pass.
\item \textbf{name}: "GiveAndGo"
\item \textbf{subtasks}: [
\texttt{Pass(A, B)},
\texttt{MoveToOpenSpace(A)},
\texttt{Pass(B, A)}
]
\item \textbf{preconditions(Pass(A, B))}: \texttt{agent(A).has\_ball == True}
\item \textbf{termination(Pass(A, B))}: \texttt{agent(B).has\_ball == True}
\item \textbf{preconditions(MoveToOpenSpace(A))}: \texttt{agent(B).has\_ball == True} (Starts after pass)
\item \textbf{termination(MoveToOpenSpace(A))}: \texttt{agent(A).is\_in\_open\_space == True}
\item \textbf{get\_intrinsic\_reward()}: $+1.0$ for each sub-task completion.
\end{itemize}

\end{document}